\documentclass{article}
\usepackage{csquotes,comment}
\usepackage{arxiv}
\usepackage[utf8]{inputenc} 
\usepackage[T1]{fontenc}    
\usepackage{hyperref}       
\usepackage{url}            
\usepackage{booktabs}       
\usepackage{amsmath,amsfonts}       
\usepackage{nicefrac}       
\usepackage{microtype}      
\usepackage{lipsum}		
\usepackage{graphicx}
\usepackage{doi}
\usepackage{multirow}
\usepackage{algorithmic, algorithm}

\title{Optimizing Neural Network Weights using Nature-Inspired Algorithms }

\author{Wael Korani \\
Computer Science Department \\
University of Regina\\
Regina, SK, Canada \\
\texttt{wmk182@uregina.ca}
\And Malek Mouhoub \\
Computer Science Department \\
University of Regina\\
Regina, SK, Canada \\
\texttt{mouhoubm@uregina.ca}
\And Samira Sadaoui \\
Computer Science Department \\
University of Regina\\
Regina, SK, Canada \\
\texttt{sadaouis@uregina.ca}}

\begin{document}
\maketitle

\begin{abstract}
This study aims to optimize Deep Feedforward Neural Networks (DFNNs) training using nature-inspired optimization algorithms, such as PSO, MTO, and its variant called MTOCL. We show how these algorithms efficiently update the weights of DFNNs when learning from data.  We evaluate the performance of DFNN fused with optimization algorithms using three Wisconsin breast cancer datasets, Original, Diagnostic, and Prognosis, under different experimental scenarios. The empirical analysis demonstrates that MTOCL is the most performing in most scenarios across the three datasets. Also, MTOCL is comparable to past weight optimization algorithms for the original dataset, and superior for the other datasets, especially for the challenging Prognosic dataset.
\end{abstract}


\section{Introduction}
\label{sec:introduction}

Medical diagnostic decision support systems (MDDSSs), which play a significant role in today's medical technology, keep growing considerably with the electronic and data extraction supports~\cite{josefson1997computers}. The ultimate purpose of MDDSSs is to assist physicians in the complex task of medical diagnosis. The cornerstone of MDDSSs is classification models that detect disease by learning from the characteristics of patient records. Several approaches were adopted to develop MDDSSs, raging from standard machine learning to deep learning. Among these approaches, researchers explored micro-array breast cancer datasets using various Artificial Neural Network (ANN) architectures, as discussed in the related work section. The Original, Diagnostic, and Prognostic Wisconsin Breast Cancer datasets (WBCDs) represent the most reliable labeled datasets that were adopted to assess different ANN models along with their weigh optimization algorithms~\cite{mangasarian1990cancer}. However, the vast majority of past literature employed the original WBCD due to its simplicity in low dimensionality and did not explore more challenging datasets to further validate the developed detection models. Across the three datasets, individual ANNs and hybrid ANNs were both investigated. Consequently, we conduct our research on breast cancer detection and employ the same datasets to compare our present work with past studies appropriately.

\smallskip

In this work, we adopt a robust Deep Feed-forward Neural Network (DFNN) to represent complex, non-linear features of breast cancer data in a better way. Nevertheless, the learning process of DFNN classifiers and its related parameters tuning is still a challenge. Thus, numerous studies were conducted to tackle the optimization of the weights of ANNs. evolutionary optimization algorithms, such as \cite{Bidar2018a, Bidar2018b, Bidar2020},  are considered reliable alternatives to optimize the DFNN weights, which motivated us to explore new meta-heuristic algorithms. In a previous study \cite{korani2020breast}, we have developed a diagnostic breast cancer framework based on three nature-inspired methods to optimize the weights of a DFNN: PSO, MTO, and MTOCL. PSO is a swarm intelligent algorithm that possesses high capabilities for difficult optimization problems. In \cite{korani2019mother}, we have introduced for the first time the variant of MTO called Mother Tree Optimization with Climate Change (MTOCL). In \cite{korani2020breast}, MTOCL returned a 100\% precision for the original WBCD, which is superior to DFNN+PSO (98.5\%) and DFNN+MTO (97.1\%). However, the authors did not report the accuracy values, so a comparison with past studies was not possible.  Also, they evaluated the three optimization algorithms with only one dataset (the simple original WBCD) and without any cross-validation (CV) and other pre-processing technique.

\smallskip

This present study is an extension of the work conducted in \cite{korani2020breast}. In this regard, we have improved the three meta-heuristic algorithms, PSO, MTO, and MTOCL, in terms of time and memory cost. Moreover, we investigate the extended algorithms using the three WBCDs, first without CV, to compare our works with past literature, then with a 10-fold CV.  In the case of Prognostic WBCD, we utilize two famous pre-processing techniques to improve the DFNN performance: the data sampling method SMOTE and the feature extraction technique PCA. We adopt SMOTE to overcome the class imbalance issue in the dataset because classifiers are influenced more by the majority class. Consequently, the minority class tends to be misclassified \cite{Anowar2020}. This issue is serious in the medical domain because the minority class (a patient has a disease) is the class of interest. We use PCA to reduce the data dimensionality and maybe increase the model accuracy. For a fair comparison with past literature, we need to perform four training scenarios of the DFNN:  (a) without CV and PCA, (b) with CV but no PCA, (c) with PCA but no CV, and (d) with CV and PCA. Actually, for each dataset, we carry out twelve training sessions of DFNN fused with the optimizers,  and with a total of 36 sessions.  However,  not all the scenarios were conducted in the literature for a given dataset. Also, no studies dealt with the imbalanced learning problem of the Prognostic WBCD.

\medskip

We structure the paper as follows. Section 2 reviews a large number of work conducted on the three WBCDs using ANN-based approaches. Section 3 presents the DFNN optimization framework and explains the workflow between the weight optimizer and the DFNN. Section 4 discusses the behavior of the three weight optimization algorithms. Section 5 exposes the three WBCDs and the preprocessing operations, including data sampling and feature extraction. Section 6 evaluates the three meta-heuristic algorithms on the three WBCDs in different experimental settings and compares the classification performance results with past literature. Section 7 summarizes our findings.

\section{Related Work}
\subsection{Studies on Original WBCD}
\label{sec:studies1}
\subsubsection{Neural Networks.}
\cite{setiono2000generating} introduced a BCD algorithm based on the quasi-Newton method and incremental hidden layers. The method is used to speed up the hidden layers' construction process. Then, for an expanded network, a set of weights is computed using the same quasi-Newton method. Through the original WBCD, the authors compared the predictive accuracy of 95.0\% with ~\cite{bennett1990neural}; the latter used back-propagation and achieved 94.9\% accuracy . Then, they proposed a pruning algorithm to remove redundant connections in the ANN and cluster the pruned hidden unit values.  They implemented the pruning algorithm for a three-layer feed-forward network, and attained more than 95\% accuracy on both training and testing data.
\medskip

\cite{west2000model} proposed a Self-Organizing Map (SOM) model where its topological properties, such as variable reduction, parameter determination and adequacy assessment for clinical measures, are used as indicators for an ideal accuracy level. The authors implemented four ANNs, Multi-Layer Perceptron (MLP), Mixture Of Experts (MOS), General Regression Neural Network (GRNN) and Radial Basis Function (RBF), and tested them on the original WBCD. They utilized the regular back-propagation algorithm (with a learning rate of 0.3 and a momentum of 0.4) for four hidden layers. Based on 10-fold cross validation (CV), the experiments showed that accuracy varies with the network architecture. The best accuracy of 97.04\% was obtained with RBF, and the three other architectures returned a high accuracy ranging from 95.72\%  to 96.29\%.
\medskip

\cite{arulampalam2001application} developed the Shunting Inhibitory Artificial Neural Network (SIANN), a biologically inspired network built on psycho-physics, speech, and perception phenomena called shunting networks. The main idea of SIANN is to build a powerful ANN with non-linear decision surfaces inspired by the non-linearity of shunting inhibition. The authors used one hidden layer with 6, 9, and 12 nodes. They evaluated four different optimization algorithms: gradient descent with an adaptive learning rate (GDA), Levenberg-Marquadt (LM), a hybrid version of direct solution (DS) and GDA called DS-GDA, and Quadratic Neural Network (QNN). SIANN attained 100\% accuracy in just few runs when evaluated on the original WBCD, and 83\% on Pima Indian Diabetes dataset. The authors then compared SIANN to MLPs, and concluded that SIANN achieved three runs (in case of LM and GDA algorithms) to six runs (in case of QNN algorithm) out of 10 runs with 100\% accuracy (no errors). However, MLPs provided the same accuracy with a maximum of three runs.
\medskip

\cite{kiyan2004breast} compared several statistical neural network structures, such as statistical ANN, GRNN, RBF and Probabilistic Neural Network (PNN) using the original WBCD. The weights of these structures are optimized using regular back-propagation algorithm. The experiments showed that the performance varies according to the statistical ANN structure. The authors recommended that the best structure is GRNN, because it achieved the highest accuracy rate of 98.8\%. The rest of the tested structures also returned relatively a high accuracy ranging from 95.74\% to 97.0\%. In general, the authors suggested that statistical neural networks could be effectively used as a BCD model.
\medskip

\cite{azmi2010breast} developed a BCD system using DFNN with regular back-propagation.  Based on the original WBCD, the model produced more cost-effective and easy-to-use classification capability. The authors evaluated the performance of several models (one hidden layer with different numbers of nodes: 4, 5, 6, and 7).  The best model achieved an accuracy of 96.63\%, with seven nodes. The rest of the models produced an accuracy above 94\%. The authors compared their results with other classifiers, and their best model is superior. Later on, \cite{pawar2013breast} evaluated a back-propagation NN model (BPNN), and compared their work with ~\cite{azmi2010breast} using a higher number of nodes (one hidden layer with different node counts: 7, 8, and 9). Increasing the number of nodes to nine improved the performance to 99\%.
\medskip

\medskip
~\cite{bevilacqua2005hybrid} adopted Genetic Algorithm (GA) as a tool to find the best topology of ANNs for the original WBCD. To search for the best solution, the authors varied the number of hidden layers from one to four, the potential max node numbers are 32, 16, 16, 16, and 8, and the activation functions are Pure linear, Tansig and Logsig. The best solution found is when the number of hidden layers is 1: a) the number of nodes per layer is 25 and the activation function is LogSig, or 2) the number of nodes is 1 and the activation function is Purelin. Although the dataset comes with only nine features, the authors applied PCA to decrease the dimensionality, and obtained 100\% accuracy, with outperform other studies.

\medskip
\cite{janghel2010breast} developed a hybrid classifier called Symbiotic-Adaptive Neuro-Evolution (SANE). The authors used a genetic operator on a population of nodes to cooperate and build a functioning neural network. Then, they compared SANE to other hybrid structures defined in previous studies: modular neural network, ensemble ANN, fixed architecture evolutionary neural network (F-ENN), and variable ENN. SANE attained an accuracy of 97.88\% and outperforms the other models that produced a performance of above 94.61\% and below 95.95\%.

\medskip
\cite{marcano2011wbcd} introduced the Artificial Metaplasticity Multilayer Perceptron (AMMLP) inspired by the biological metaplasticity property of neurons as well as the Shannon's information theory.  The AMMLP idea is to update the weights of less frequent activation to achieve a more efficient training. The results showed that AMMLP returned 99.26\% accuracy, which is higher than the classical back-propagation algorithm and other algorithms, such as C4.5 (using CV) with 94.74\%, RAIC with 95.0\%, LDA with 96.8\%, NEFCLASS with 95.06, Fuzzy-GA1 with 97.36\%, Neuro-rule 2a with 98.10\%, LSA machine with 98.8\% , SFC with 95.57\%, SVM (99.54\%), LS-SVM (98.53\%), LLS (96\%), SVM-CFS (99.51\%), AR+NN (97.40\%), and CFW (99.5\%). However, the majority of the compared studies were not ANN-based models.

\subsubsection{Hybrid Neural Networks.}
~\cite{karabatak2009expert} proposed a diagnostic breast cancer system by combining Association Rules (AR) and ANNs. The idea here was to introduce a dimension reduction method that may help enhancing the classification accuracy. The authors used AR to decrease the dimensionality from nine to \textit{four} of the original WBCD; the new inputs are: Uniformity of Cell Size (UCSize), Normal Nucleoli (NN), Bare Nuclei (BN), and Mitoses. Based on 3-fold CV, the accuracy using only ANN but without AR is 95.2\%, and with AR is 95.6\%.

\medskip
\cite{ubeyli2005mixture} introduced a modular neural network, called Mixture of Experts (ME). The author used Expectation-Maximization (EM) to train the ME. The proposed model was evaluated with the original WBCD, and returned 98.85\% accuracy, which is higher than a stand-alone neural network. Later in~\cite{ubeyli2009adaptive}, the same author implemented an adaptive neuro-fuzzy inference system (ANFIS) combined with both ANN, which has an adaptive capability, and the quantitative approach fuzzy logic. The hybrid system produced 99.08\% accuracy, which is pretty higher than the first model.
\medskip

\cite{ashraf2010information} proposed a model by combining an Adaptive Network based Fuzzy Inference System (ANFIS) and Information Gain (GI) technique. GI was implemented to reduce the number of inputs and ANFIS to map inputs to outputs. In fact, GI gives more details about the dataset and dedicates for each attribute a specific rank that helps in selecting the most significant attributes. Based on the original WBCD, the authors compared their results with several past studies. ANFIS achieved 98.24\% accuracy, which is higher than previous performance results.
\medskip

\cite{khosravi2011breast} proposed a hybrid system combining three modules: fuzzy feature extraction, training, and MLP. The training module merges the Bees Algorithm (BA) with the regular back-propagation algorithm called BA-BP to leverage both global and local searches. Based on the original WBCD, the authors compared BP, BA, and BA-PA algorithms with and without fuzzy feature extraction. BA-PA with feature selection is the most effective module as it achieved the highest performance of 97.83\%. Then, the authors compared two different classifiers: RBF and probabilistic neural networks (PNNs) with and without fuzzy feature selection.  RBF with feature selection is the best classifier with an accuracy of 95.15\%. They also compared three different hybrid training systems: Genetic Algorithm BP (GA-BP), Imperialist Competitive Algorithm BP (ICA-BP), and BA-BP. The best hybrid system is BA-BP as it attained 97.83\%. Lastly, they reported the best result of 99.42\% for the proposed method called FCONN, however, did not state anywhere what FCONN stands for.


\subsection{Studies on Several WBCDs}
\label{sec:studies2}
\cite{chunekar2009approach} examined the efficiency of the Jordan Elman Neural Network (JENN), a Recurrent Neural Network (RNN). RNN has a single hidden layer that possesses feedback connections from the hidden output nodes to the inputs of the network. However, JENN has feedback connections from the output nodes to the inputs. JENN was trained using several factors: quick prop delta bar delta, momentum, Leven beg Marqua, and conjugate gradient method. It achieved high accuracy of 98.75\% on original WBCD, 98.25\% on diagnostic WBCD, and only 70.725\% on prognostic WBCD.

\medskip

\cite{salama2012breast} combined different classifiers to produce a suitable meta-classifier that provides high accuracy for the three WBCDs using PCA as the preprocessing step. First, the authors implemented five classifiers on the original WBCD: Naive Bayes (NB), Decision Tree (J48), Sequential Minimal Optimization (SMO), Instance-Based for K-Nearest neighbor (IBK) and MLP. SMO is the best classifier that returned 96.9957\% of accuracy. Then, the authors developed ensemble models: NB+SMO, MLP+SMO, J48+SMO, and IBK+SMO, and showed that NB+SMO, MLP+SMO and IBK+SMO  have the same performance that the baseline SMO. The accuracy level increased when the three classifiers were combined together, SMO+IBK+NB, to reach 97.1388\%, and when four classifiers were combined, SMO+IBK+NB+J48, to reach 97.2818\%. For the diagnostic WBCD, the best model returned 97.7153\% using SMO only. Lastly, the meta-classifier SMO+J48+MLP+IBK provided only 77.3196\% when tested on the prognostic WBCD. The authors concluded that using SMO only, SMO+MLP and SMO+IBK are the most performing.

\medskip

Although fully connected neural networks (FCNNs) received much attention for their generalization power, \cite{belciug2010partially} adopted a partial connected neural network (PCNN) to speed up the training process and reduce computation resources. The suggested model is inspired by how the human brain works: if a signal is processed, only a few neurons are activated, and the rest is inhibited. PCNN, consisting of fully connected neurons, is generated by dividing the main problem into smaller sub-problems/modules that are fully connected.  The authors conducted an extensive experiment using original, diagnostic and prognostic WBCDs, and Ljubljana Recurrence Breast Cancer (LRBC).  With 10-fold CV, the results of PCNN using original WBCD is 94.21\%, diagnostic WBCD is 81.08\%, and prognostic WBCD is only 71.2\%. In case of prognostic dataset, PCNN achieved better performance than FCNN. The results of FCNN using original WBCD is 95.22\%, diagnostic WBCD is 81.39\%, and prognostic WBC is 60.21\%.

\subsection{Discussion}
As we can observe, the vast majority of breast cancer detection models did not conduct enough experiments to prove their models' effectiveness because they investigated only the simple original WBCD. These models achieved reasonably high accuracy when tested with this dataset, but a low accuracy for the prognostic dataset, ranging from 60\% to 70\%. The studies did not provide reliable neural network structures that could attain high performance for different WBCDs. Furthermore, most studies just used the standard back-propagation algorithm to optimize the weights and biases of the ANNs where the estimated error at the output layer is propagated backward to the hidden and input layers to update the weights and minimize the loss function. In the present work, we select three nature-inspired optimization methods: PSO (proposed in 1995), MTO (proposed in 2019), and MTOCL {(a variant of MTO proposed in 2019}. The latter can escape from local optima using climate change events and achieve better results as demonstrated in~\cite{korani2019mother}.

\section{DFNN Optimization Framework}
\begin{figure}[!ht]
\begin{center}
	\includegraphics[width=0.70\linewidth]{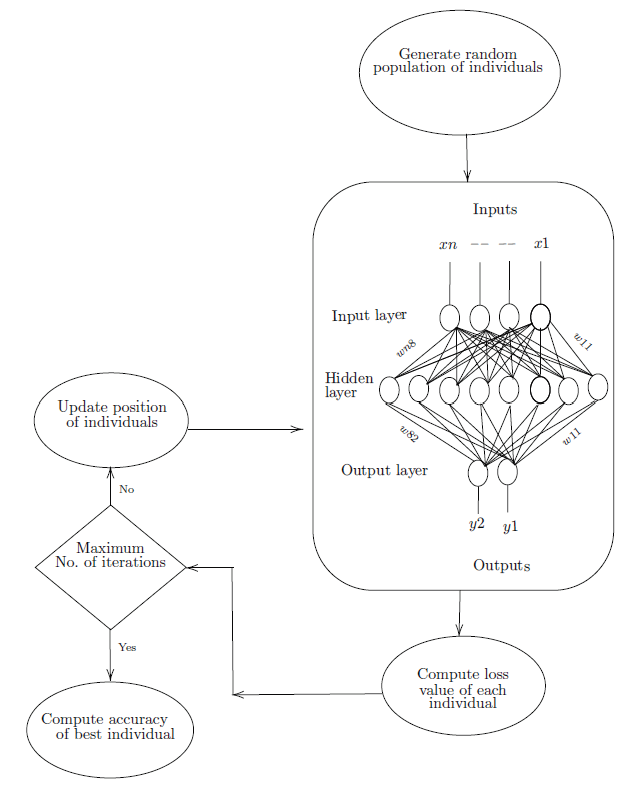}
\caption{Optimizing the Weights of Deep Feed-Forward Neural Networks}
   \label{model1}
    \end{center}
\end{figure}

Our proposed optimization framework consists of two primarily parts: the weight optimizer and the DFNN architecture, as depicted in Figure~\ref{model1}. The DFNN is a fully connected multi-layer perceptron that consists of the input layer, a certain number of hidden layers, and an output layer. The input layer has a number of nodes equals to the dimensions of the training dataset. The input data are forward propagated through the network to produce the corresponding labels. To properly compare the three optimization algorithms, we adopt the same structure for the hidden layers across all the experiment settings. In fact, we select the number of hidden layers/nodes after conducting preliminary experiments on the three training datasets. The output layer has two output nodes as we are dealing with binary classification. We choose the Relu activation function for all the hidden layers, Sigmoid function for the output layer, and the Root Mean Square Error (RMSE) as the loss function to be minimized by the meta-heuristic optimization algorithms. We limit the weight values to the range of [5, -5].

\medskip
The weight optimizer has one of the three optimization algorithms, PSO, MTO, or MTOCL. In the first iteration, the algorithm generates a random population of agents; an agent represents a vector of the weights of the entire DFNN architecture. The first generated agents (candidate solutions) are sent to the DFNN, where each agent in the population produces a particular neural network model based on its own weights. Next, the loss function of each agent is computed, and sent to the optimizer. The latter updates all the agents (weight vector) based on the strategy of the implemented optimization algorithm. The updated population is sent back to the DFNN, and this process keeps iterating until a stopping criteria is reached; in our case the number of iterations. In the last iteration, the agent that attains the lowest loss value is chosen  as the best solution i.e., the optimal weights of the DFNN. The best agent is then used to compute the DFNN accuracy in our experiments on the one hand, and to classify new patient records on the other hand.

\section{Optimization Algorithms for DFNN}
Optimizing the weights of DFNN classifiers is a crucial stage to increase the learning efficiency. However, it is considered a challenging task in the area of optimization. In this section, we describe the behavior of  PSO, MTO, and MTOCL according to the framework of Figure 1.

\subsection{Particle Swarm Optimization (PSO)}
In~\cite{eberhart1995new}, Eberhart and Kennedy introduced the main idea of PSO, which builds on the movement of a flock of birds. The population of PSO is a group of particles ($P_{n}$); each one denotes a bird. Each particle in the population updates its position based on two primary components: \enquote{cognitive} and  \enquote{social}. The cognitive component is the local vector toward the local best position, and is calculated with $[P_{id}^{k} - x_{id}^{k}]$, where $P_{id}^{k}$ is the local best position of particle $i$ at iteration $k$, and $x_{id}$ its current position at iteration $k$. The social component is the global vector toward the global best position, and is calculated with $[P_{gd}^{k} - x_{id}^{k}]$, where $P_{gd}^{k}$ is the global best position at iteration $k$. In our work, we actually implement the canonical PSO variant, because it has the constriction factor that helps in controlling the convergence properties of the particles~\cite{clerc2002particle}. The position of each particle is updated by adding the velocity vector as follows:

\begin{equation}
 x_{id}^{k+1} = x_{id}^{k} + v_{id}^{k+1}
 \label{xeq}
\end{equation}
where $v_{id}$ is the velocity of particle $i$ at iteration $k+1$. The velocity vector of a particle is updated by the following equation:
\begin{equation}
 v_{id}^{k+1} =  \chi (v_{id}^{k} + c_{1} \times r_{1} [P_{id}^{k} - x_{id}^{k}] + c_{2} \times r_{2} [P_{gd}^{k} - x_{id}^{k}])
 \label{veq}
\end{equation}
where $\chi$ is the constriction factor, $v_{id}^{k}$ the velocity of particle $i$ at iteration $k$,  $c_{1}$ and $c_{2}$ the acceleration coefficients, and $r_{1}$ and $r_{2}$ the random distribution numbers in the range [0, 1]. Algorithm~\ref{PSO} presents the main steps of the Canonical PSO algorithm. In our specific optimization problem, the particle position represents the weights of the entire neural network, and the fitness value of each particle the DFNN loss value. In addition, the stopping criteria is represented by the number of iterations ($Iters$).

\begin{algorithm}[ht]
\caption{Canonical PSO}
\label{PSO}
\begin{algorithmic}
\STATE {\bf Inputs}: Iters, n, c1, c2, r1, r2, $\chi$
\STATE {\bf Outputs}: $P_{gbest}$
\STATE Generate random positions and velocities ($P_n$ and $V_{n}$)
\WHILE{Iters not reached }
\FOR{particle $i$ = 1 to $n$}
    \STATE Update the velocity $V_{i}$ and the position $P_{i}$ using Eq.~\ref{veq} and Eq. \ref{xeq}
    \STATE Update the fitness value {$fi$}
    \IF {$f_{i}$ $<$ local best fitness ($f_{lbest}$)}
       \STATE  $P_{lbest} = P_{i}$
    \ENDIF
    \IF{$f_{i}$ $<$ global best fitness ($f_{gbest}$)}
        \STATE  $P_{gbest} = P_{i}$
    \ENDIF
\ENDFOR
\ENDWHILE
\STATE {\bf return} $P_{gbest}$
\end{algorithmic}
\end{algorithm}

\subsection{Mother Tree Optimization (MTO)}
\begin{figure}[!ht]
\begin{center}
	\includegraphics[width=0.95\linewidth]{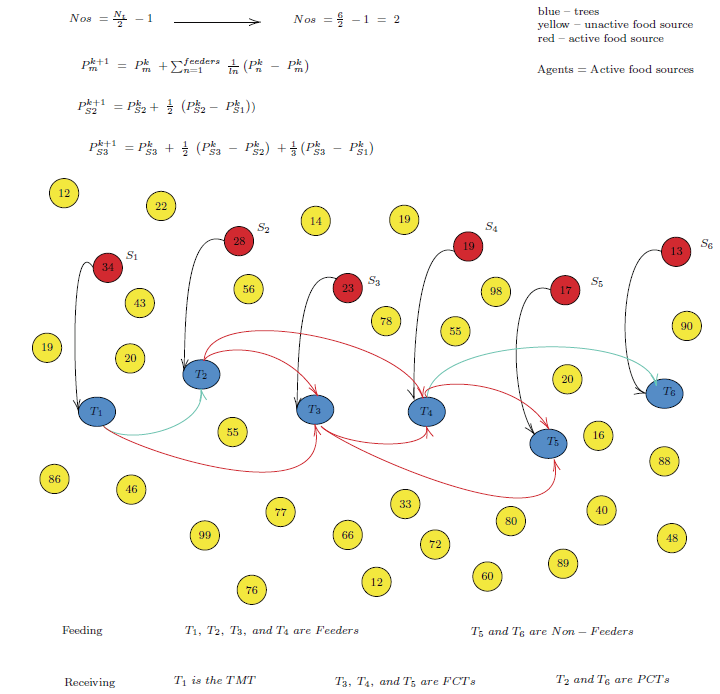}
\caption{A group of active food sources and trees in a search space}
    \label{AFS}
    \end{center}
\end{figure}

In~\cite{korani2019mother}, we introduced the main idea of the MTO method, which is inspired by the feeding behavior of Douglas Fir trees with essential support of Mycorrhizal Fungi Network (MFN) that facilitates the nutrient transferring between plants of same or different species. The search space has unlimited number of food sources (FSs) and six trees (as an example) as depicted in Figure~\ref{AFS}. The trees, marked blue, select some of the FSs to be Active Food Sources (AFS that are marked red). In each iteration, AFSs are replaced by AFSs in the search space. Figure~\ref{AFS} is an example with a certain number of food sources (yellow and red) and a number of nutrients assigned to each AFS. In each iteration, the number of AFSs is equal to the number of trees. The population size is denoted by $N_{T}$. In our particular optimization problem, an AFS denotes the weights of the whole DFNN, the fitness value of an AFS is the loss value of the DFNN, and the stopping criteria is the number of kin recognition signals (Krs).

Agents in the MTO population are sorted in descending order of their fitness values, so that the tree $T_1$ has the highest nutrient value and we call it the Top Mother Tree (TMT). Each agent in the population feeds or receives nutriens. Figure~\ref{AFS} illustrates an example of an MTO population where the feeding and receiving process is indicated using the arrows. According to the feeding process, the population is divided into feeders and non-feeders, and all feeders will feed an offspring $N_{os}$. The number of agents in the feeder and non-feeders groups is computed as follows.

\begin{equation}
  	N_{os} = \frac{N_{T}}{2} - 1 \\
    N_{T} = N_{Frs} + N_{NFrs}\\
    N_{Frs} = \frac{N_{T}}{2} + 1
\end{equation}

According to the receiving process, the population is divided into three main groups: TMT, Fully Connected Trees (FCTs), and Partially Connected Trees (PCTs). Each group has its own protocol to update the position of its members. Firstly, the TMT is the one agent that has the highest number of nutrients, and it just receives nutrients from a random food source. The FCT group has $\text{N}_{\text{FCTs}}$ agents, and the PCT group has $\text{N}_{\text{PCTs}}$ agents. According to the above equations, the size of PCTs and FCTs groups are computed as follows.

\begin{equation}
  \begin{gathered}
    N_{PCTs}  = N_{T} - 4 \\
    N_{FCTs} = 3 \\
    N_{T} = N_{FCTs} + N_{PCTs} + 1
  \end{gathered}
\end{equation}

The TMT introduces two levels of exploitation to find a better solution close to its current position. The position is updated for the first level as follows.

\begin{equation}
   P_{1}(x_{k+1}) = P_{1}(x_{k}) + \delta  R(d)
  \label{rs}
\end{equation}

\begin{equation}
R(d) = 2 \ (round \  ( rand( d , 1 ) )- 1 ) \ rand(d , 1)
\label{rs1}
\end{equation}

where $\delta$ denotes the root signal, $R(d)$ a random vector that can change by different seed numbers. $k$ is the current iteration number. 
Once the TMT position is updated according to the above equation, then it moves in a random direction with a small $\Delta$ step size.
\begin{equation}
\small
P_{1}(x_{k+1}) = P_{1}(x_{k}) +  \Delta R(d)
\label{ams}
\end{equation}
 $\Delta$ is the mycorrhizal fungi network step size. The values of $\delta$ and $\Delta$ have significant effect on the performance of the MTO algorithm.

The PCTs has two groups: First Partially Connected Trees (FPCTs) and Last partially Connected Trees (LPCTs). The FPCTs group has $(\frac{\text{N}_{\text{T}}}{2} -2)$ agents. The members of this group are in the range [2, $(\frac{\text{N}_{\text{T}}}{2}-1$)]. A member of FPCT group updates its position as follows:

\begin{equation}
\small
P_{n}(x_{k+1}) = P_{n}(x_{k}) + \sum _{i = 1}^{n-1}\frac{1}{n-i+1}(P_{i}(x_{k})-P_{n}(x_{k}))
\label{fpcts}
\end{equation}
where $P_{n}(x_{k})$ and $P_{n}(x_{k+1})$ respectively denote the current and the updated positions of the feeders of agent n.  If the defense mechanism is used, the equation will be updated as follows. 
\begin{equation}
\small
P_{n}(x_{k+1}) = P_{n}(x_{k}) +  \phi R(d),
\label{defense}
\end{equation}
Here, $\phi$ is a slight deviation from the current position~\cite{korani2019mother}, which will enable or disable the defense mechanism. LPCTs start at candidate solution ranked $\frac{\text{N}_{\text{T}}}{2}+3$ to the end of the population. The updated position is computed as follows. 

\begin{equation}
\small
P_{n}(x_{k+1}) = P_{n}(x_{k}) + \sum _{i = n-N_{os}}^{N_{T}-N_{os}}\frac{1}{n-i+1}(P_{i}(x_{k})-P_{n}(x_{k}))
\label{lpcts}
\end{equation}

The members of the FCT group is located in the range [$(\frac{\text{N}_{\text{T}}}{2})$ : $(\frac{\text{N}_{\text{T}}}{2}+2)$]. The updated position is computed below:

\begin{equation}
\small
P_{n}(x_{k+1}) = P_{n}(x_{k}) + \sum _{i = n-N_{os}}^{n-1}\frac{1}{n-i+1}(P_{i}(x_{k})-P_{n}(x_{k}))
\label{fcts}
\end{equation}

\subsection{Mother Tree Optimization with Climate Change (MTOCL)}
In~\cite{korani2019mother}, we introduced a new variant of MTO, called MTOCL.  MTOCL introduces two additional operations to prevent agents in the population from being stuck in local minima: elimination and distortion as shown in Algorithm~\ref{MTOCL}. In the elimination operation, the agents with the lowest fitness are removed and replaced by new random agents in the search space. In the distortion operation, the rest of the population is deviated from its current position.The elimination percentage (El) is recommended to be 20\% based on experiments we conducted on several complex optimization problems~\cite{korani2019mother}. In the latter study, MTOCL outperforms MTO and both algorithms were superior to PSO.

\begin{algorithm}[ht]
\caption{$ \text{MTO and MTOCL    }$}
\label{MTOCL}
\begin{algorithmic}
\STATE {\bf Inputs:}
\STATE $ N_{T} \text{: Size of the population}$
\STATE $ K_{rs} \text{: number of iterations (kin recognition signals)}$
\STATE $ Cl \text{: number of climate change events (operations, where 0 is for MTO)}$
\STATE $ El \text{: elimination percentage} $
\STATE {{\bf Outputs:}  {$P_{best}$}}
\STATE $\text{Generate a population of } N_{T} \text{ agents} (P_{1},\dots,P_{N_T})$
\STATE $\text{Evaluate the fitness values for all the agents }$
\STATE $\text{Sort all the agents in descending order of their fitness and store them in }S$

	\LOOP
		\FOR{$k_{rs} = 1$ to $K_{rs}$}
		    \STATE $\text{Use equations (\ref{rs})--(\ref{fcts}) to update the position of each agent in }S $
			\STATE $\text{Evaluate the fitness of the updated agents  } $
			\STATE $\text{Sort the agents in descending order and store them in }S$
		\ENDFOR
		
		\IF{$Cl = 0$}    \STATE $BREAK$
		\ELSE 
			\STATE $\text{Select and distort the best agents in S according to El}$
			\STATE $\text{and replace the remaining ones with new random agents}$
		
			\STATE $Cl = Cl - 1 $
		\ENDIF
	 \ENDLOOP  
	
\RETURN $P_{best}$
\end{algorithmic}
\vspace{\baselineskip}
\end{algorithm}

\section{Wisconsin Breast Cancer Datasets}
There are two options regarding breast cancer data: they are either based on gene expression or the cell'\ s characteristics.  In this paper, we test the effectiveness of the SI optimization algorithms on three WBCDs, Original, Diagnostic, and Prognostic; all of them depend on the tumor cell's characteristics. The original WBCD possesses nine numerical predictors: Sample Number, Clump Thickness, Uniformity of Cell Size, Uniformity of Cell Shape, Marginal Adhesion, Single Epithelial Cell Size, Bare Nuclei, Bland Chromatin, Normal Nucleoli, and Mitoses. The target class is Benign (B) or Malignant (M).

\smallskip

The Diagnostic WBCD has 30 classification attributes~\cite{street1993nuclear} with the two target labels of B and M. Ten real-valued features were measured for each cell nucleus: Radius, Texture, Smoothness,  Concavity, Perimeter, Area,  Compactness,Concave Points, Symmetry, Fractal Dimension~\cite{wolberg1995computerized}. Then, the mean, standard error and worst/largest values were computed for each of the ten attributes, resulting in 30 predictors for this dataset.

\smallskip

The Prognostis WBCD comes with 32 prediction attributes~\cite{street1993nuclear} along with two outputs: Recurrent (R) or Non-Recurrent (NR). In addition to the 30 attributes of the Diagnostic WBCD, two attributes were added: Time and Tumor Size.  This dataset has five missing records that we eliminate.  This dataset is challenging to learn from because the count of samples is relatively small compared to the dimensionality. In this situation, the performance may be affected due to the overfitting issue. As mentioned in~\cite{verleysen2005curse}, for robust learning, the number of samples must grow exponentially with the number of attributes, which means this dataset necessitates at least $10^{32}$ samples.

\begin{table}[h]
    \centering
    \begin{tabular}{|c|c|c|c|}
    \hline
       &  {\bf Original} & {\bf Diagnostic} & {\bf Prognostic} \\ \hline
      No. Instances   & 699 & 569 & 198 \\ \hline
      Dimensionality   &  9 & 30   & 32 \\ \hline
      Labels & B and M & B and M & R and NR \\ \hline
      Class Distribution    & 2:1   & 2:1   & 3:1   \\
        Ratio   &  (458 vs. 241) &   (357 vs. 212) &  (151 vs. 47) \\
      \hline
    \end{tabular}
    \caption{Wisconsin Breast Cancer Datasets}
    \label{Description}
\end{table}

Table~\ref{Description} presents some statistics about the three WBCDs. Since each WBCD possesses features with different scales, we then normalize all the data to the range of [0, 1] to improve the predictive power.  Moreover, we analyze the class distribution in the data, and as exposed in Table~\ref{Description}, the first two datasets have a good class imbalance ratio (negative to positive instances) of 2:1. However, the Prognostic dataset has a ratio of 3:1 that may lower the accuracy. Thus, we adopt the popular over-sampling method SMOTE to re-balance the dataset with a ratio of 2:1 to be consistent with the two other datasets and increase the number of observations to train the DFNN classifier better and reduce overfitting.  Another reliable option is to reduce the data dimensionality by applying a feature extraction method, and the most popular one is the Principal Component Analysis (PCA).  The latter merges similar features into new features by preserving the original feature properties ~\cite{Raj2019}. Still, we need to search for the optimal number of new features that will lead to the highest accuracy for this dataset.

\section{Evaluation and Comparison}
We assess the classification performance of our DFNN fused with each weight optimization algorithm, PCO, MTO, and MTOCL, using the three WBCDs. Additionally, we compare our work with existing studies discussed in the related work section. As observed, these studies adopted different ANN architectures (individual ANNs or combined with other learning algorithms) and conducted different types of experiments using or not CV (10 folds) and dimensionality reduction. For a fair comparison, we need to perform four training scenarios of the DFNN:  (a) without CV and PCA, (b) with CV but no PCA, (c) with PCA and no CV, and (d) with CV and PCA.  We may note that no studies dealt with the imbalanced learning problem of the Prognostic WBCD. Before training the DFNN on the WBCDs, we first tune the hyper-parameters of the three weight optimization methods.

\subsection{Parameter Tuning}
We perform many experiments to determine the optimal hyper-parameter values of the three optimization algorithms for the three WBCDs, as presented in Table \ref{settings}.

\begin{small}
\begin{table}[!ht]
\caption {\textsc{Hyper-Parameter Tuning for the Three Optimization Algorithms}}
\label{settings}
\begin{center}
\begin{tabular}{ c | c  }
\hline
Algorithm 		&  Parameter Setting \\
\hline
{MTO}         &   Root signal $\delta = 1.0 $ \\
  or              &   MFN signal $\Delta = 0.3 $\\
{MTOCL}  &   Small deviation $\phi = 1 $ \\
				& $N_{T}$ = 20 \\
				& Cl = 5 \\
				& EL = 20\% \\
				& Iters = 500 \\
\hline
PSO			    &  $\chi  = 0.72984 $\\
				& $c_{1} = 2.02$ and $c_{1} = 2.02$  \\
				& $r_{1}$ and $r_{2}$ are random \\
				& n = 20 \\
				& Iters = 500\\
				
\hline
\end{tabular}
\end{center}
\label{para}
\end{table}
\end{small}

\subsection{Original WBCD}
\begin{figure*}[h!]
 \centering
   \medskip
    \centering
    \includegraphics[width=.40\linewidth]{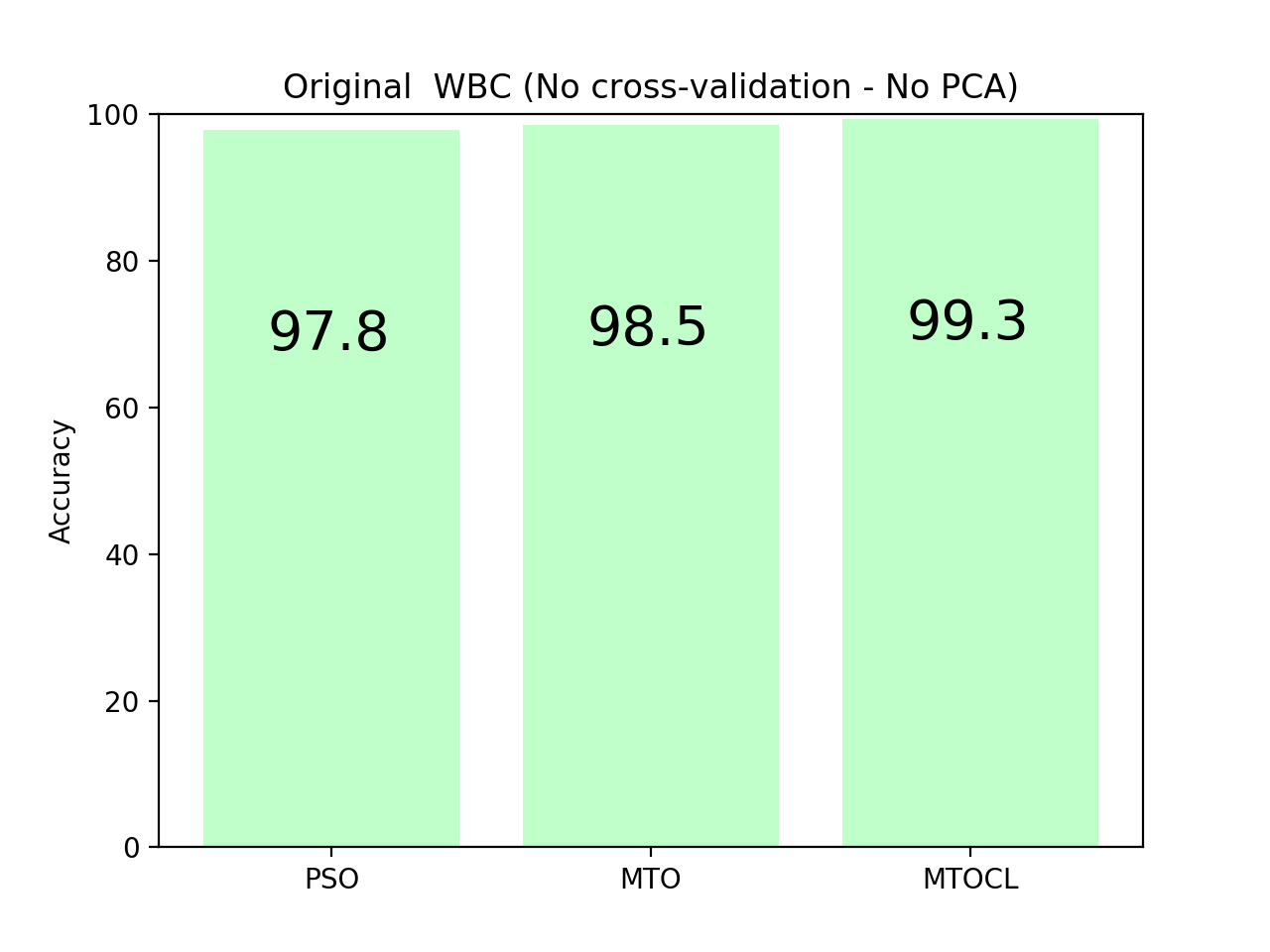}
  \quad
    \centering
    \includegraphics[width=.40\linewidth]{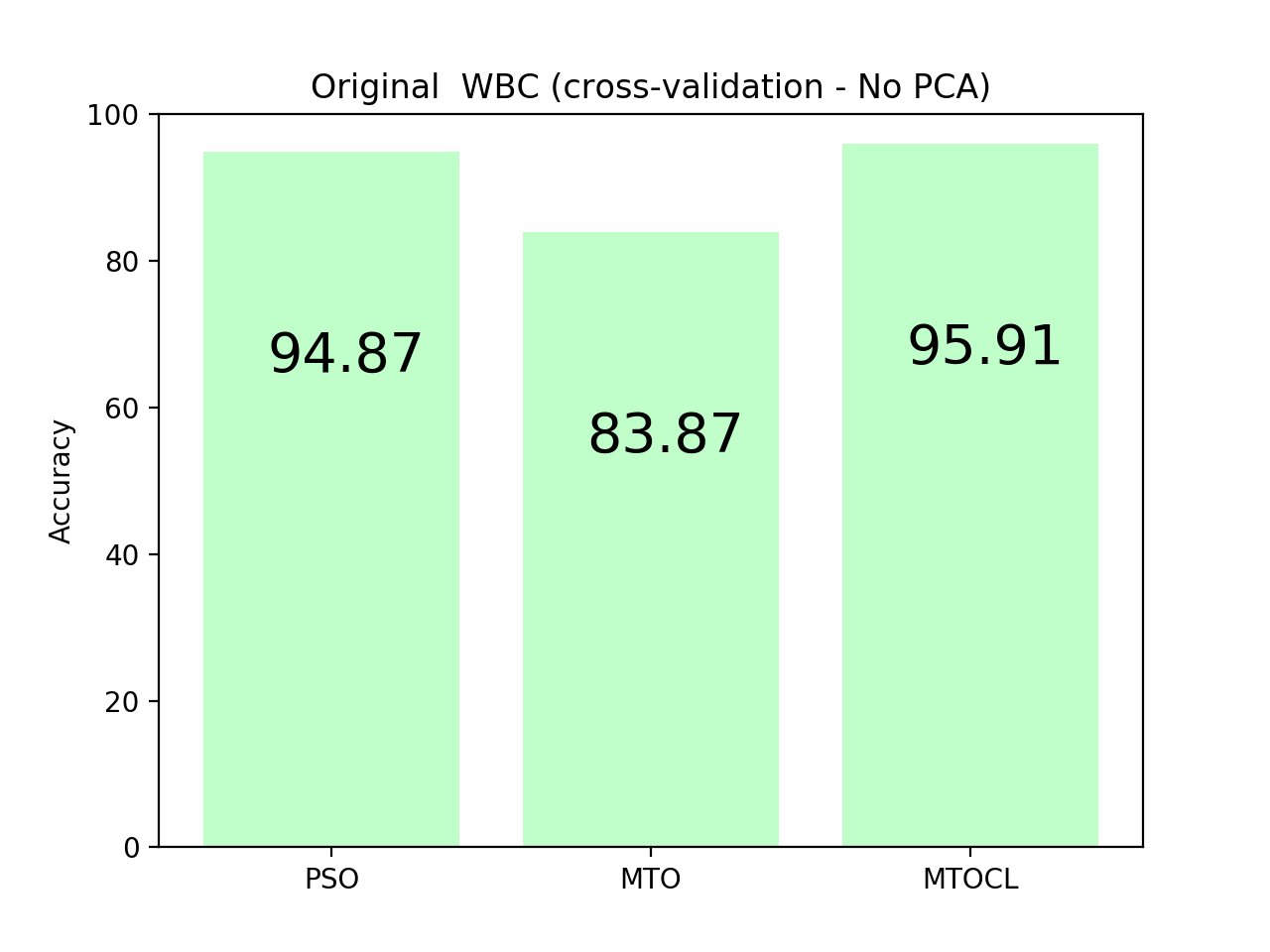}

    \centering
    \includegraphics[width=.40\linewidth]{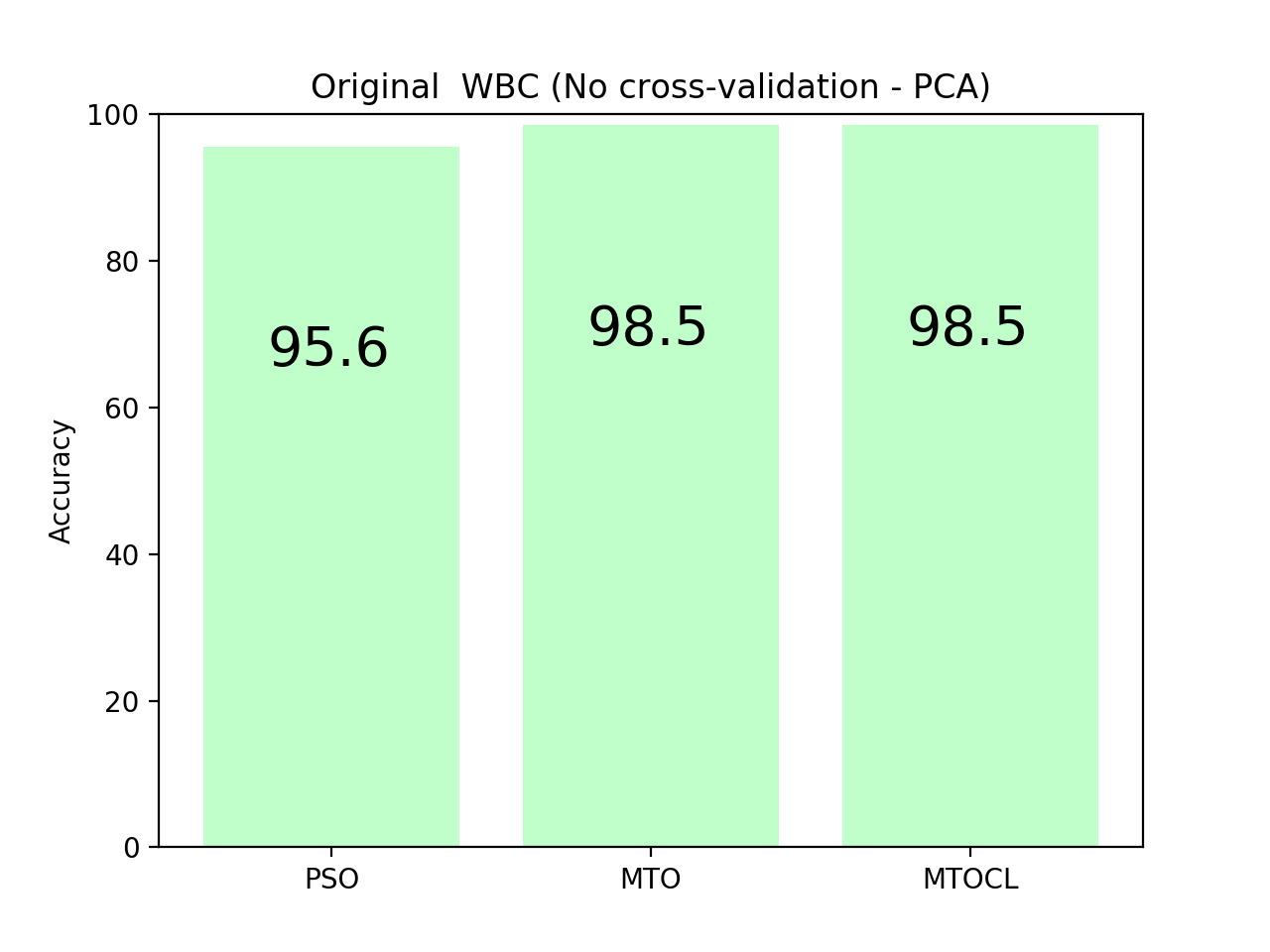}
  \quad
    \centering
    \includegraphics[width=.40\linewidth]{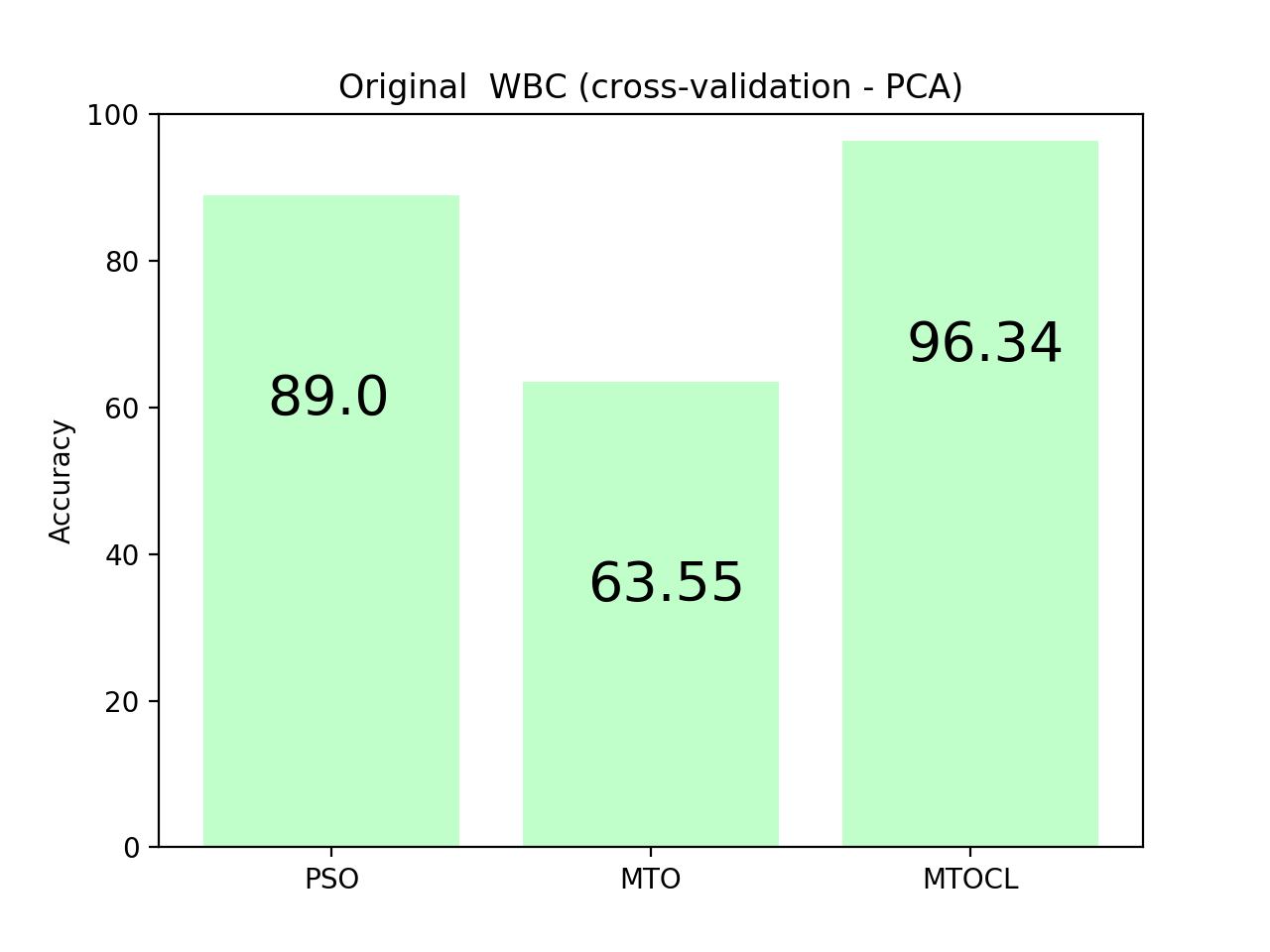}
  \caption{Accuracy of Original WBCD: (a) No CV \& No PCA , (b) CV \& No PCA , (c) No CV \& PCA, (d) CV \& PCA}
  \label{OriginalPerf}
\end{figure*}

The first part of Table~\ref{tab:oldResults} reports the performance of previous breast cancer diagnostic models reviewed in the related work section using only the original WBCD.  Figure~\ref{OriginalPerf} exposes the accuracy results of our optimization algorithms on this dataset.

\begin{table}[]
\begin{center}
\begin{tabular}{|c|c|c|c|}
\hline
{\bf Original}  & {\bf Accuracy} \\ \hline
\cite{bennett1990neural}      &  94.9\%                                                                \\ \hline
\cite{setiono2000generating}    & 98.0\%                                                                \\ \hline
\cite{west2000model}                  &  97.04\%                                                               \\ \hline
\cite{arulampalam2001application}        & 100.0\%                                                               \\ \hline
\cite{kiyan2004breast}                 &   98.8\%                                                                \\ \hline
\cite{bevilacqua2005hybrid}           &    100\%                                                                 \\ \hline
\cite{karabatak2009expert}            &    95.6\%                                                                \\ \hline
\cite{ubeyli2005mixture}             &    98.85\%                                                               \\ \hline
\cite{ubeyli2009adaptive}             &    99.08\%                                                               \\ \hline
\cite{azmi2010breast}                  &   96.63\%                                                               \\ \hline
\cite{pawar2013breast}                  & 99.0\%                                                                \\ \hline
\cite{janghel2010breast}                 &97.88\%                                                               \\ \hline
\cite{ashraf2010information}              & 98.24\%                                                               \\ \hline
\cite{marcano2011wbcd}                     &99.26\%                                                               \\ \hline
\cite{khosravi2011breast}                   & 99.42\%                                                               \\  \hline \hline
{\bf Original, Diagnostic } &  {\bf Accuracy} \\ 
{\bf   and Prognostic} &  \\ 
\hline
\cite{chunekar2009approach}       &  98.75\%, 98.25\%,  70.725\%  \\ \hline

\cite{salama2012breast}           &  97.2818\%, 97.7153\%, 77.3196\%                                       \\ \hline
\cite{belciug2010partially}       &  94.21\%, 81.08\%, 71.2\%                                              \\ \hline
\end{tabular}
\end{center}
\caption{Accuracy of past breast cancer detection models}
\label{tab:oldResults}
\end{table}

\begin{enumerate}
\setlength{\itemsep}{2pt}
\item  First experiment type: In Figure~\ref{OriginalPerf}(a), MOTCL optimizer achieved the highest accuracy of 99.3\%, but it is still comparable to the two other algorithms. With the same training setting of ANNs,~\cite{arulampalam2001application} produced 100\% in some runs. \vspace*{0.2cm}

\item  Second experiment type: In Figure~\ref{OriginalPerf}(b), MTOCL is again the most performing, and the accuracy is still relatively high after learning with 10-fold CV. MTOCL is comparable to PSO but outperforms MTO with an increase of 12\%. This gap is significant in medical diagnosis.  With this setting, MTOCL is superior to~\cite{belciug2010partially} that obtained an accuracy of 94.21\%.  \vspace*{0.2cm}

\item  Third experiment type: In Figure~\ref{OriginalPerf}(c), we first reduce the dimensionality from nine to four inputs as done in~\cite{bevilacqua2005hybrid} that produced 100\% accuracy after applying PCA. In our case, the best accuracy is 98.5\% with MTOCL and MTO, which is actually lower than using the whole set of features. Another study \cite{khosravi2011breast} lowered the dimensionality from nine to two using a clustering algorithm, and attained a performance of 99.42\%. \vspace*{0.2cm}

\item  Fourth experiment type: In Figure~\ref{OriginalPerf}(d), MOTCL again attains the highest accuracy, with an increase of 7.34\% compared to PSO and 32.79\% compared to MTO. No existing studies used this training setting.
\end{enumerate}

\subsection{Diagnostic WBCD}
The last three records in Table~\ref{tab:oldResults} is about studies that used the diagnostic WBCD. ~\cite{chunekar2009approach} returned 98.25\%, which is the best among the three studies.

\begin{figure*}[h!]
  \centering
    \centering
    \includegraphics[width=.40\linewidth]{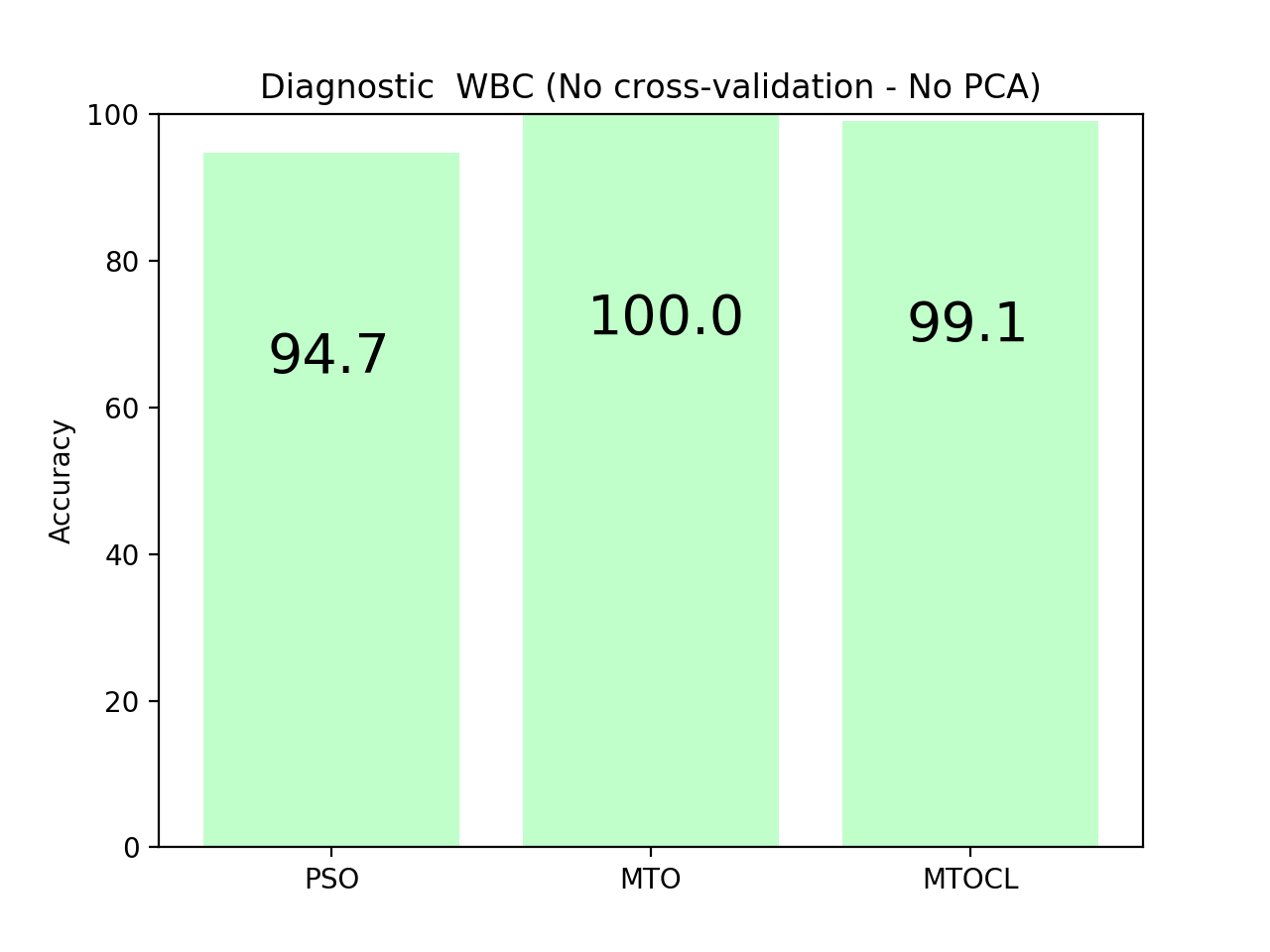}
    \quad
    \centering
    \includegraphics[width=.40\linewidth]{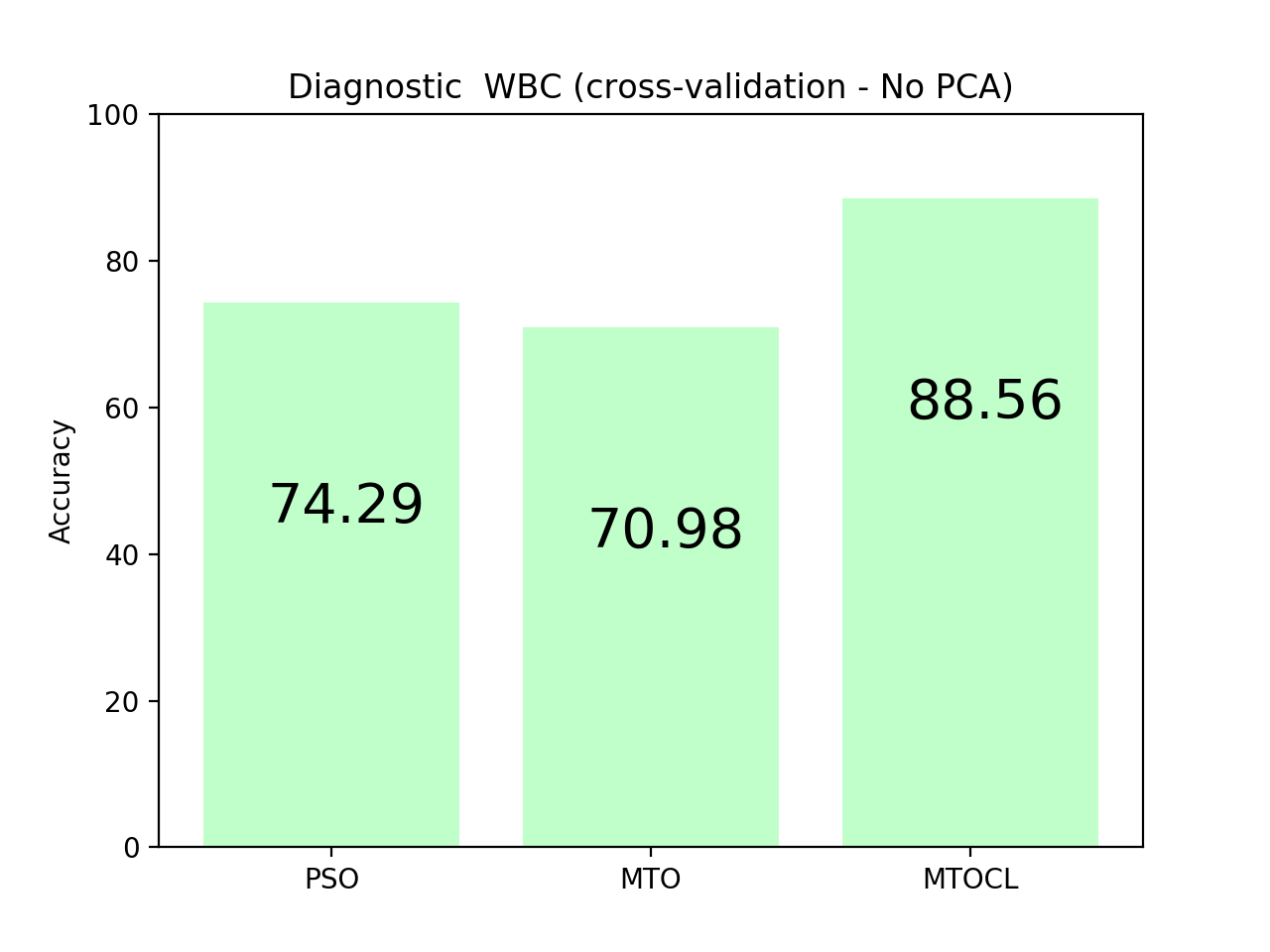}
 \quad
    \centering
    \includegraphics[width=.40\linewidth]{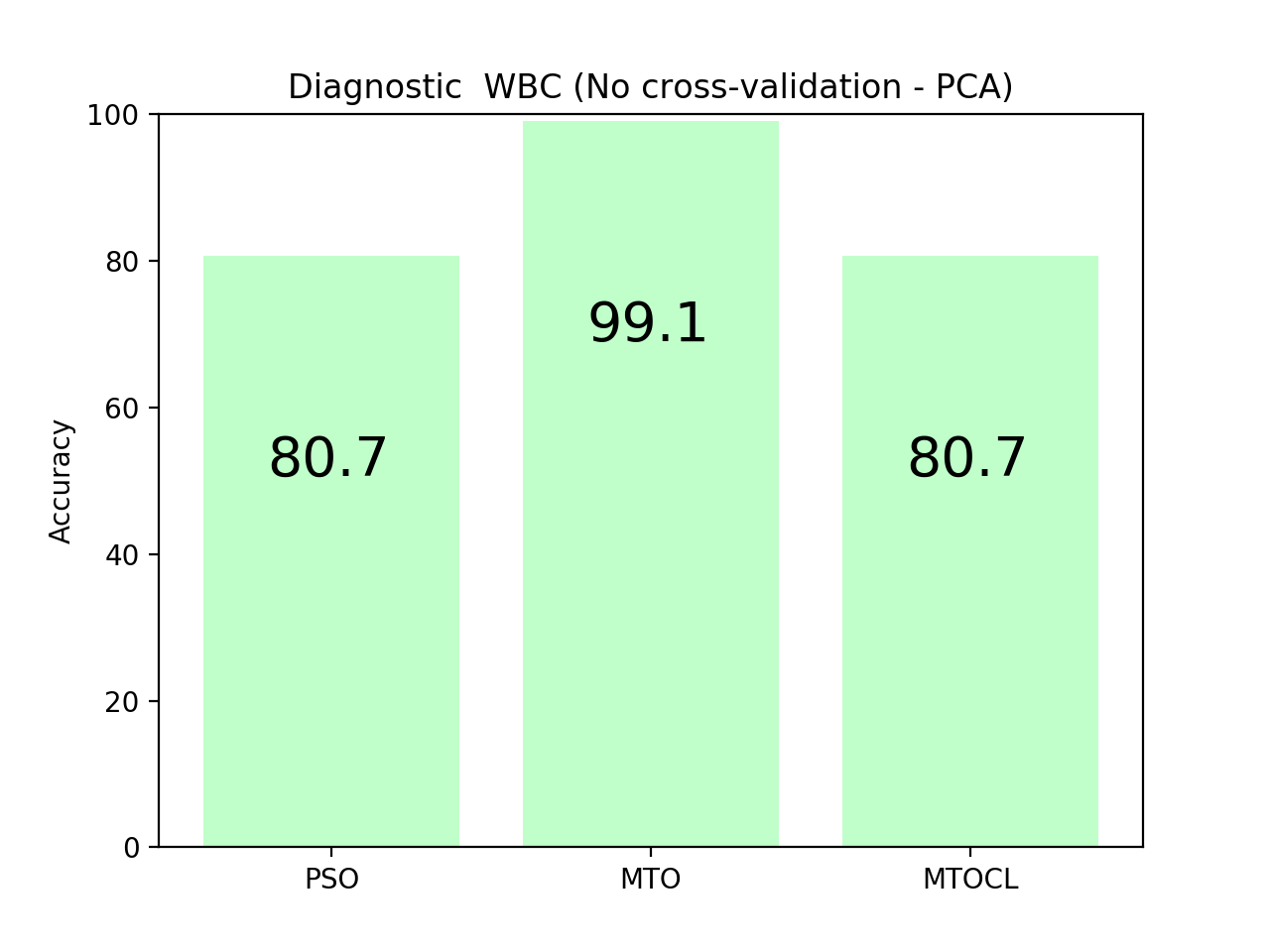}
    \quad
    \centering
    \includegraphics[width=.40\linewidth]{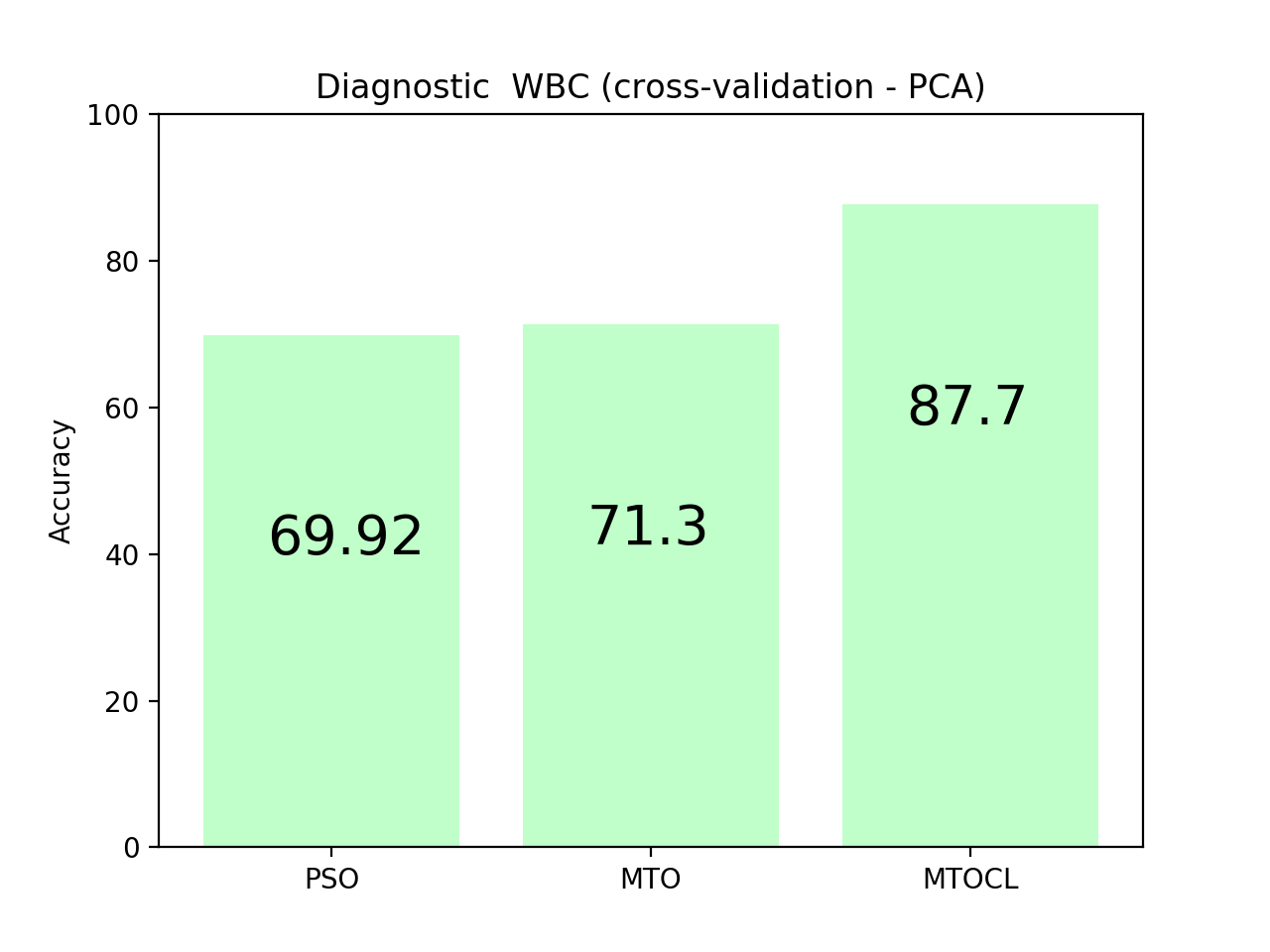}
 \caption{Accuracy of Diagnostic WBCD: (a) No CV \& No PCA (b) CV \& No PCA (c) No CV \& PCA, (d) CV \& PCA}
 \label{DiaPer}
\end{figure*}

\begin{enumerate}
\item First experiment type: In Figure~\ref{DiaPer}(a), MTO returned an outcome of 100\%, which is higher than the best results achieved in~\cite{chunekar2009approach} and \cite{belciug2010partially} using the same setting. MTO is comparable to MTOCL, but is superior to PSO with a gap of 5.3\%.   \vspace*{0.2cm}
\item Second experiment type: In Figure~\ref{DiaPer}(b), MTOCL is the most performing with a significant increase of 14.27\% compared to PSO and 17.58\% compared to MTO. MTOCL performance is better that the one reported in~\cite{belciug2010partially}.
\vspace*{0.2cm}
\item Third experiment type: In Figure~\ref{DiaPer}(c), MTO is the best classifier, and PSO and MTOCL provided the same performance. The authors in~\cite{belciug2010partially} claimed that using PCA for their classifiers did not improve the performance.
\vspace*{0.2cm}
\item Fourth experiment type: In Figure~\ref{DiaPer}(d), MTOCL did much better than the other algorithms, and outperformed MTO by 16.4\% and PSO by 17.78\%. These gaps are significantly high in breast cancer detection. Previous work did not carry out this setting for this dataset.
\end{enumerate}

\subsection{Prognostic WBCD}
\begin{figure}[ht]
  \centering
    \centering
    \includegraphics[width=.35\linewidth]{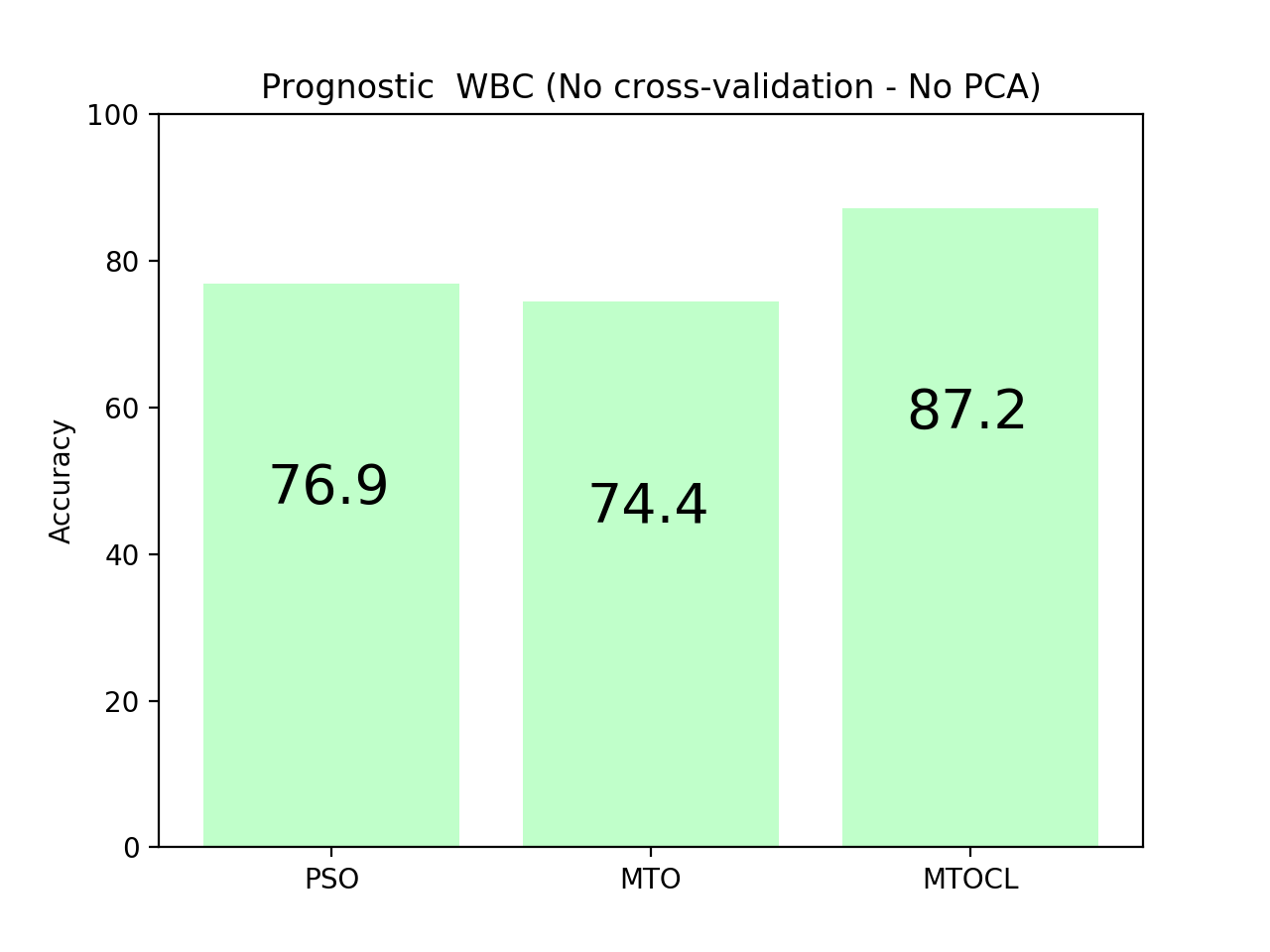}
  \quad
    \centering
    \includegraphics[width=.35\linewidth]{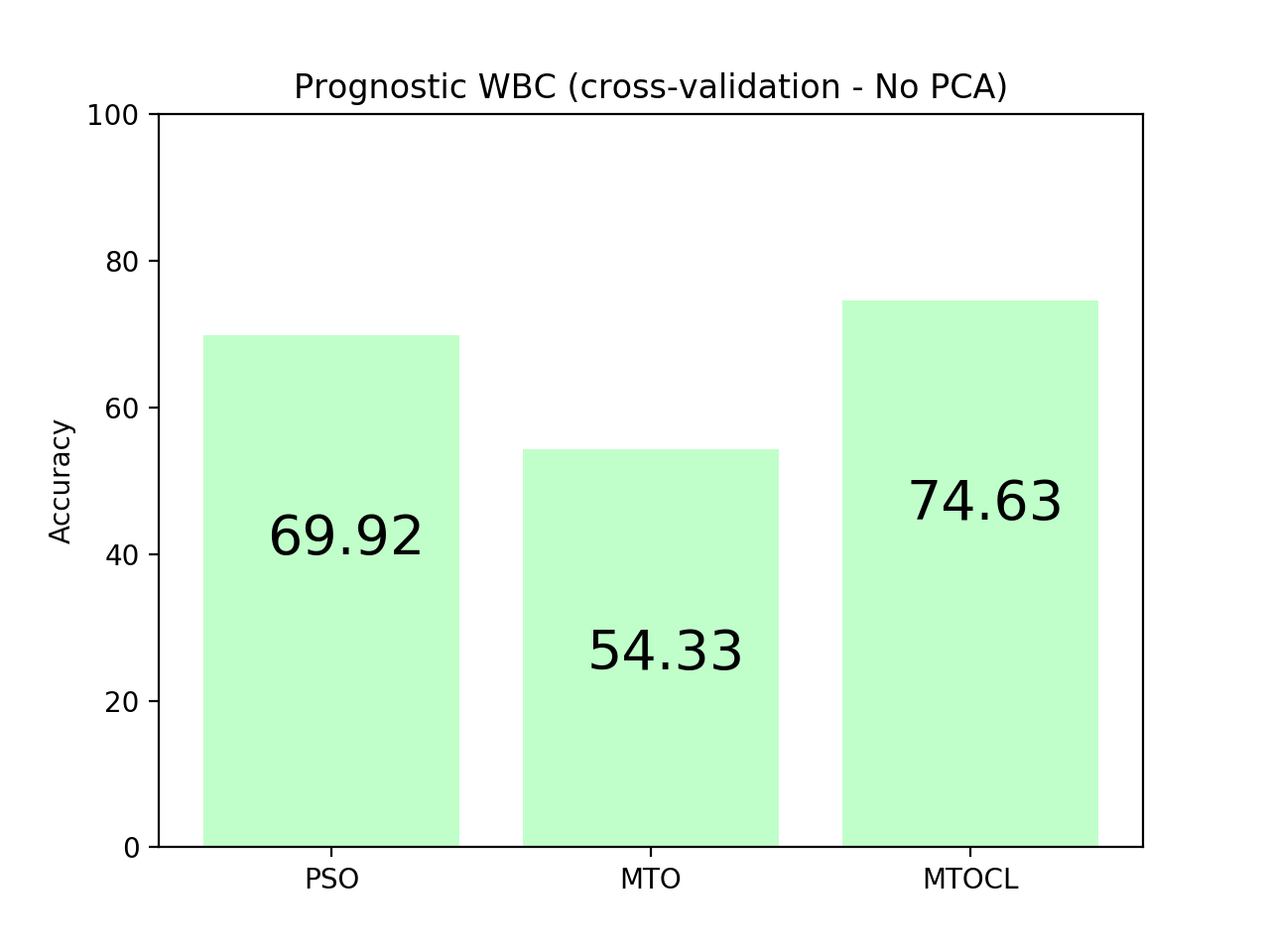}
    \centering
    \includegraphics[width=.35\linewidth]{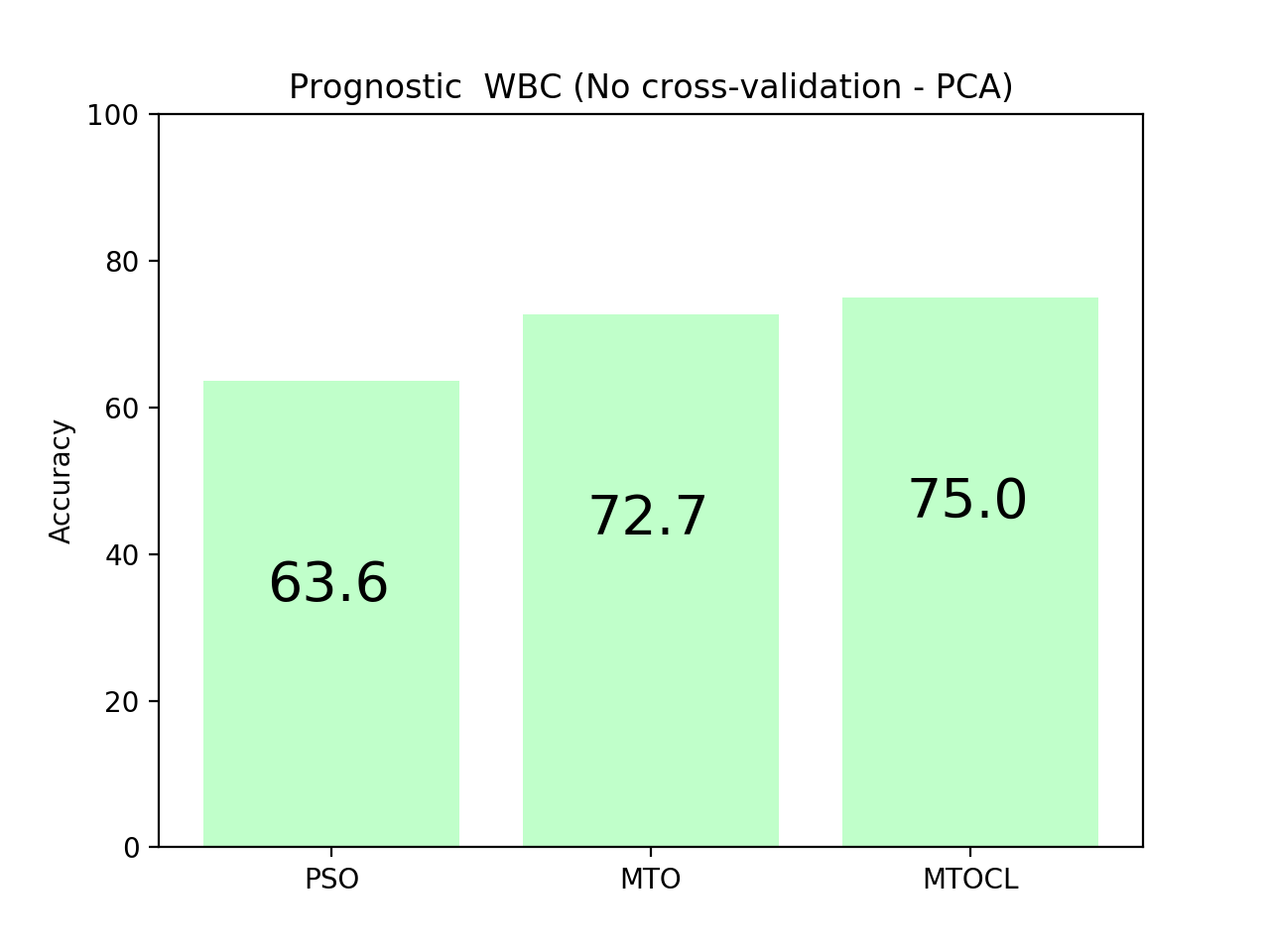}
     \quad
    \centering
    \includegraphics[width=.35\linewidth]{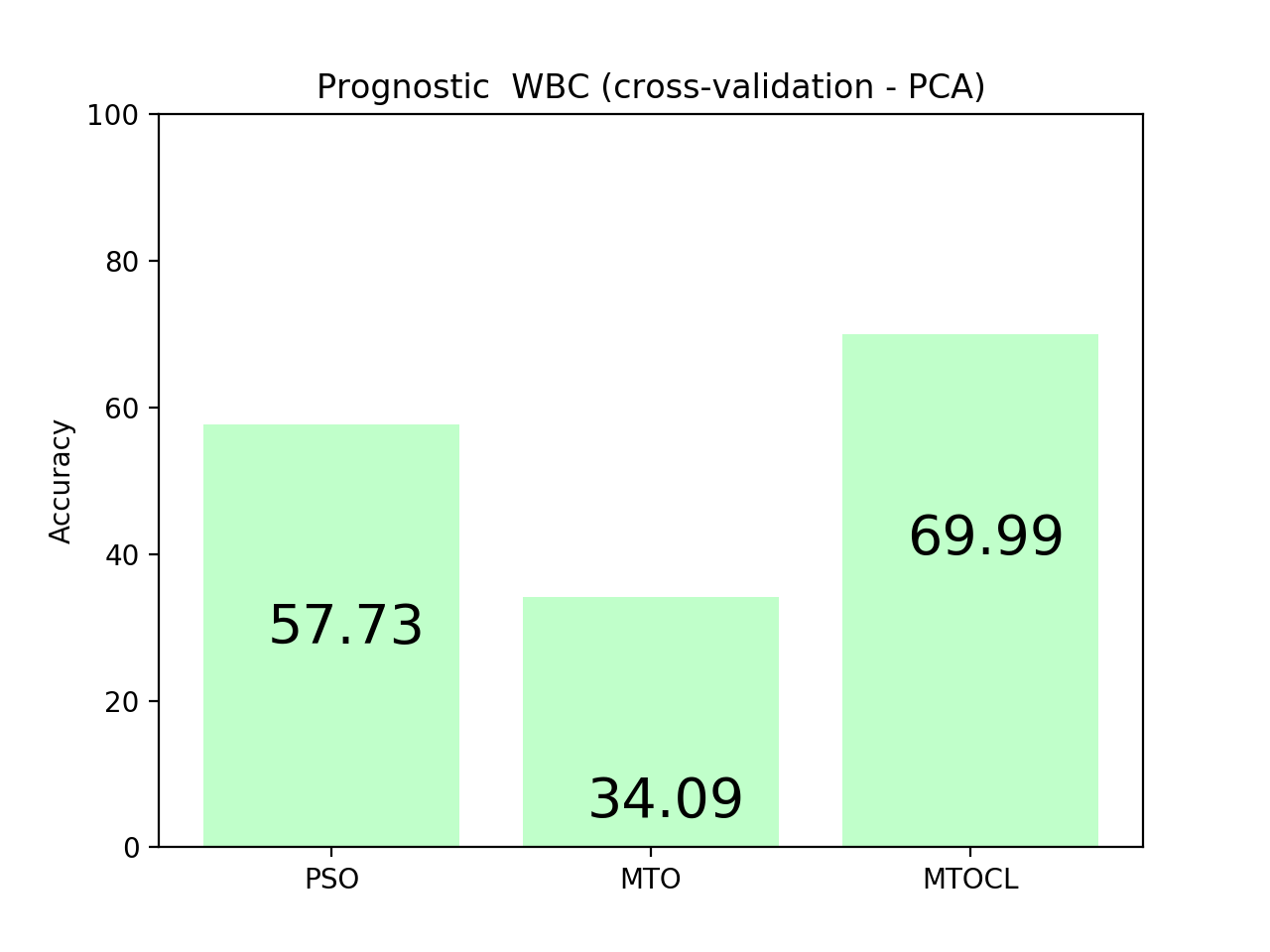}
    \caption{Accuracy of Prognostic WBCD: (a) No CV \& No PCA , (b) CV \& No PCA , (c) No CV \& PCA, (d) CV  \& PCA}
    \label{proPer}
\end{figure}

\begin{enumerate}
\item First experiment type: In Figure~\ref{proPer}(a), MTOCL outperforms PSO and MTO with a difference of 10.3\% and 12.8\% respectively.  The MTOCL accuracy of 87.2/\% is greater than other last studies of Table~\ref{tab:oldResults} where the highest performance is 71.2\%~\cite{belciug2010partially}. We note that \cite{belciug2010partially} and \cite{chunekar2009approach} utilized the same setting, it means no cross validations.

\vspace*{0.2cm}

\item Second experiment type: Again MTOCL is superior to PSO and MTO. We may note that MTO did not do well for this setting. MTOCL outcome is pretty close to~\cite{salama2012breast} who attained 77.3196\%~\cite{salama2012breast}. The work~\cite{salama2012breast} adopted an ensemble of classifiers: SMO+J48+MLP+IBK and received an accuracy of 77.3196\%. However, our model is much simpler and returned a close accuracy.

\vspace*{0.2cm}

\item Third experiment type: We firstly apply PCA to reduce the dimensionality from 32 to 8 features. After extensive experiments, we determine the optimal number of principal components to be eight. In Figure~\ref{OriginalPerf}(c), the best accuracy is 75.0\% using MTOCL. No studies conducted this setting for this dataset.

\vspace*{0.2cm}

\item Fourth experiment type: Again MTOCL is the optimal optimization method. It outperformed MTO with an increase of 35.0\% for MTO and 12.26\% for PSO. We may note that MTO did poorly in this context. Prior research did not perform within this setting.

\end{enumerate}

\subsection{Discussion}
All the classification results are summarized in Table~\ref{tab:results}. Regarding the original dataset, MTOCL is the clear winner across all the settings. MTOCL was able to provide a very high accuracy of 99.3\% without conducting any CV and feature extraction. For the diagnostic dataset, MTOCL is the most performing algorithm in two settings (when using CV), and is comparable to the best outcome in another setting. Thus, we select MTOCL as the preferable optimizer for the second dataset, as it returned a performance of 99.1\% without CV and feature extraction. For the prognostic dataset, MTOCL is by far the optimal optimizer across all the settings with the promising accuracy of 87.7\% again without CV and PCA.  According to all the experimental results, we can observe that this dataset is challenging to learn from since except for one case, the optimizers produced low accuracy. The latter is that the dataset has a few samples but many features.  Despite this challenge, MTOCL was able to train our DFNN efficiently on the one hand, and defeat past studies on the other hand. So, we can conclude that MTOCL is superior to PSO and MTO for all the WBCDs. Additionally, these datasets do not require any CV nor dimensionality reduction due to the low number of samples and features.

{\renewcommand{\arraystretch}{1.15}}%
\begin{table*}[h!]
\centering

\begin{tabular}{|c@{\qquad}|c@{\qquad}|c@{\qquad}|c@{\qquad}|c@{\qquad}|}

\hline

{\bf WBCD} & {\bf Training}   & {\bf PSO } & {\bf MTO } & {\bf MTOCL }  \\
  &    & {\bf  + DFNN} & {\bf  + DFNN} & {\bf  + DFNN}  \\
\hline

\multirow{4}{*}{Original} & No-CV and No-PCA  &  97.8   & 98.5 &  \bf 99.3 \\
\cline{2-5}
                                & CV and No-PCA  & 94.87  & 83.87    &  \bf 95.91
                                \\ \cline{2-5}
                                & No-CV and PCA & 95.6   & \bf 98.5 & \bf 98.5
                                \\ \cline{2-5}
                                & CV and PCA    &  89.0  &  63.55  &  \bf 96.93
                                \\ \hline

\multirow{4}{*}{Diagnostic}        & No-CV and No-PCA & 94.7 & \bf 100.0 & 99.1     \\ \cline{2-5}
                                & CV and No-PC & 74.29 & 70.98  & \bf 88.56    \\
 \cline{2-5}
                                & No-CV and PCA & 80.7    & \bf 99.1 &  80.7     \\ \cline{2-5}
                                & CV and PCA     & 69.92 &  71.3  & \bf 87.7   \\ \hline

\multirow{4}{*}{Prognostic}        & No-CV and No-PCA  & 76.9 & 74.4 & \bf 87.2    \\ \cline{2-5}
                                & CV and No-PCA &  69.92 & 54.33 & \bf 74.63
                                \\ \cline{2-5}
                                &  No-CV and PCA & 63.6  &  72.7  &  \bf 75.0
                                \\ \cline{2-5}
                                & CV and PCA &  57.73  & 34.09  &  \bf 69.99
                                \\ \hline
\end{tabular}
\caption{Accuracy of WBCDs using different combinations of CV and PCA}
\label{tab:results}
\end{table*}
\section{Conclusion}
We combined deep neural network architectures and nature-inspired optimization methods to develop a robust breast cancer detection framework. The latter will support physicians as a complementary recommendation in the diagnoses of cancer.  More precisely,  through three cancer datasets, we assessed three meta-heuristic optimization techniques.  Based on different training scenarios, the experimental analysis demonstrated that MTOCL+DFNN achieved the highest accuracy of 99.3\% for the original WBCD,  MTO+DFNN a 100\% accuracy for the diagnostic WBCD, and MTOCL+DFNN an 87.2\% for the prognostic WBCD. In most of the experimental scenarios, MTOCL outperformed the other two optimization algorithms. Also, MTOCL is comparable to past weight optimization algorithms for the original dataset and superior for the two other datasets.


\begin{thebibliography}{10}

\bibitem{josefson1997computers}
Deborah Josefson.
\newblock {Computers Beat Doctors in Interpreting ECGs}.
\newblock {\em {British Medical Journal Publishing Group}}, 1997.

\bibitem{mangasarian1990cancer}
Olvi~L Mangasarian and William~H Wolberg.
\newblock Cancer diagnosis via linear programming.
\newblock Technical report, University of Wisconsin-Madison Department of
  Computer Sciences, 1990.

\bibitem{Bidar2018a}
Mahdi Bidar, Hamidreza~Rashidy Kanan, Malek Mouhoub, and Samira Sadaoui.
\newblock {Mushroom Reproduction Optimization (MRO): A Novel Nature-Inspired
  Evolutionary Algorithm}.
\newblock In {\em {Congress on Evolutionary Computation, CEC}}, pages 1--10.
  {IEEE}, 2018.

\bibitem{Bidar2018b}
Mahdi Bidar, Malek Mouhoub, and Samira Sadaoui.
\newblock {Discrete Firefly Algorithm: A New Metaheuristic Approach for Solving
  Constraint Satisfaction Problems}.
\newblock In {\em {Congress on Evolutionary Computation, CEC}}, pages 1--8.
  {IEEE}, 2018.

\bibitem{Bidar2020}
Mahdi Bidar, Malek Mouhoub, Samira Sadaoui, and Hamidreza~Rashidy Kanan.
\newblock {A Novel Nature-Inspired Technique Based on Mushroom Reproduction for
  Constraint Solving and Optimization}.
\newblock {\em {International Journal of Computational Intelligence and
  Applications}}, 19(2):2050010:1--2050010:21, 2020.

\bibitem{korani2020breast}
Wael Korani and Malek Mouhoub.
\newblock Breast cancer diagnostic tool using deep feedforward neural network
  and mother tree optimization.
\newblock In {\em International Conference on Optimization and Learning}, pages
  229--240. Springer, 2020.

\bibitem{korani2019mother}
Wael Korani, Malek Mouhoub, and Raymond~J Spiteri.
\newblock Mother tree optimization.
\newblock In {\em 2019 IEEE International Conference on Systems, Man and
  Cybernetics (SMC)}, pages 2206--2213. IEEE, 2019.

\bibitem{Anowar2020}
Farzana Anowar and Samira Sadaoui.
\newblock {Detection of Auction Fraud in Commercial Sites}.
\newblock {\em {Journal of Theoretical and Applied Electronic Commerce
  Research, JTAER}}, 15(1):81--98, 2020.

\bibitem{setiono2000generating}
Rudy Setiono.
\newblock Generating concise and accurate classification rules for breast
  cancer diagnosis.
\newblock {\em Artificial Intelligence in medicine}, 18(3):205--219, 2000.

\bibitem{bennett1990neural}
Kristin~P Bennett and Olvi~L Mangasarian.
\newblock Neural network training via linear programming.
\newblock Technical report, University of Wisconsin-Madison Department of
  Computer Sciences, 1990.

\bibitem{west2000model}
David West and Vivian West.
\newblock Model selection for a medical diagnostic decision support system: a
  breast cancer detection case.
\newblock {\em Artificial Intelligence in medicine}, 20(3):183--204, 2000.

\bibitem{arulampalam2001application}
Ganesh Arulampalam and Abdesselam Bouzerdoum.
\newblock Application of shunting inhibitory artificial neural networks to
  medical diagnosis.
\newblock In {\em The Seventh Australian and New Zealand Intelligent
  Information Systems Conference, 2001}, pages 89--94. IEEE, 2001.

\bibitem{kiyan2004breast}
T{\"u}ba Kiyan and T{\"u}lay Yildirim.
\newblock Breast cancer diagnosis using statistical neural networks.
\newblock {\em Istanbul University-Journal of Electrical \& Electronics
  Engineering}, 4(2):1149--1153, 2004.

\bibitem{azmi2010breast}
Muhammad Sufyian Bin~Mohd Azmi and Zaihisma~Che Cob.
\newblock Breast cancer prediction based on backpropagation algorithm.
\newblock In {\em 2010 IEEE Student Conference on Research and Development
  (SCOReD)}, pages 164--168. IEEE, 2010.

\bibitem{pawar2013breast}
Punam~S Pawar and Dharmaraj~R Patil.
\newblock Breast cancer detection using neural network models.
\newblock In {\em 2013 International Conference on Communication Systems and
  Network Technologies}, pages 568--572. IEEE, 2013.

\bibitem{bevilacqua2005hybrid}
Vitoantonio Bevilacqua, Giuseppe Mastronardi, and Filippo Menolascina.
\newblock Hybrid data ananlysis methods and artificial neural network design in
  breast cancer diagnosis: Idest experience.
\newblock In {\em International Conference on Computational Intelligence for
  Modelling, Control and Automation and International Conference on Intelligent
  Agents, Web Technologies and Internet Commerce (CIMCA-IAWTIC'06)}, volume~2,
  pages 373--378. IEEE, 2005.

\bibitem{janghel2010breast}
Rekh~Ram Janghel, Anupam Shukla, Ritu Tiwari, and Rahul Kala.
\newblock Breast cancer diagnostic system using symbiotic adaptive
  neuro-evolution (sane).
\newblock In {\em 2010 International Conference of Soft Computing and Pattern
  Recognition}, pages 326--329. IEEE, 2010.

\bibitem{marcano2011wbcd}
Alexis Marcano-Cede{\~n}o, Joel Quintanilla-Dom{\'\i}nguez, and Diego Andina.
\newblock Wbcd breast cancer database classification applying artificial
  metaplasticity neural network.
\newblock {\em Expert Systems with Applications}, 38(8):9573--9579, 2011.

\bibitem{karabatak2009expert}
Murat Karabatak and M~Cevdet Ince.
\newblock An expert system for detection of breast cancer based on association
  rules and neural network.
\newblock {\em Expert systems with Applications}, 36(2):3465--3469, 2009.

\bibitem{ubeyli2005mixture}
Elif~Derya {\"U}beyli.
\newblock A mixture of experts network structure for breast cancer diagnosis.
\newblock {\em Journal of medical systems}, 29(5):569--579, 2005.

\bibitem{ubeyli2009adaptive}
Elif~Derya {\"U}beyli.
\newblock Adaptive neuro-fuzzy inference systems for automatic detection of
  breast cancer.
\newblock {\em Journal of medical systems}, 33(5):353, 2009.

\bibitem{ashraf2010information}
Muhammad Ashraf, Kim Le, and Xu~Huang.
\newblock Information gain and adaptive neuro-fuzzy inference system for breast
  cancer diagnoses.
\newblock In {\em 5th International Conference on Computer Sciences and
  Convergence Information Technology}, pages 911--915. IEEE, 2010.

\bibitem{khosravi2011breast}
Alireza Khosravi, Jalil Addeh, and Javad Ganjipour.
\newblock Breast cancer detection using ba-bp based neural networks and
  efficient features.
\newblock In {\em 2011 7th Iranian Conference on Machine Vision and Image
  Processing}, pages 1--6. IEEE, 2011.

\bibitem{chunekar2009approach}
Vaibahv~Narayan Chunekar and Hemant~P Ambulgekar.
\newblock Approach of neural network to diagnose breast cancer on three
  different data set.
\newblock In {\em 2009 International Conference on Advances in Recent
  Technologies in Communication and Computing}, pages 893--895. IEEE, 2009.

\bibitem{salama2012breast}
Gouda~I Salama, M~Abdelhalim, and Magdy Abd-elghany Zeid.
\newblock Breast cancer diagnosis on three different datasets using
  multi-classifiers.
\newblock {\em Breast Cancer (WDBC)}, 32(569):2, 2012.

\bibitem{belciug2010partially}
Smaranda Belciug and Elia El-Darzi.
\newblock A partially connected neural network-based approach with application
  to breast cancer detection and recurrence.
\newblock In {\em 2010 5th IEEE International Conference Intelligent Systems},
  pages 191--196. IEEE, 2010.

\bibitem{eberhart1995new}
Russell Eberhart and James Kennedy.
\newblock A new optimizer using particle swarm theory.
\newblock In {\em MHS'95. Proceedings of the Sixth International Symposium on
  Micro Machine and Human Science}, pages 39--43. IEEE, 1995.

\bibitem{clerc2002particle}
Maurice Clerc and James Kennedy.
\newblock The particle swarm-explosion, stability, and convergence in a
  multidimensional complex space.
\newblock {\em IEEE transactions on Evolutionary Computation}, 6(1):58--73,
  2002.

\bibitem{street1993nuclear}
W~Nick Street, William~H Wolberg, and Olvi~L Mangasarian.
\newblock Nuclear feature extraction for breast tumor diagnosis.
\newblock In {\em Biomedical image processing and biomedical visualization},
  volume 1905, pages 861--870. International Society for Optics and Photonics,
  1993.

\bibitem{wolberg1995computerized}
William~H Wolberg, W~Nick Street, Dennis~M Heisey, and Olvi~L Mangasarian.
\newblock Computerized breast cancer diagnosis and prognosis from fine-needle
  aspirates.
\newblock {\em Archives of Surgery}, 130(5):511--516, 1995.

\bibitem{verleysen2005curse}
Michel Verleysen and Damien Fran{\c{c}}ois.
\newblock The curse of dimensionality in data mining and time series
  prediction.
\newblock In {\em International work-conference on artificial neural networks},
  pages 758--770. Springer, 2005.

\bibitem{Raj2019}
Judy~T. Raj.
\newblock A beginner’s guide to dimensionality reduction in machine learning,
  2019.

\end{thebibliography}

\end{document}